# Spatial Aggregation: Theory and Applications

**Kenneth Yip**  YIP@MARTIGNY.AI.MIT.EDU
*MIT Artificial Intelligence Laboratory, 545 Technology Square*
*Cambridge, MA 02139 USA*

**Feng Zhao**  FZ@CIS.OHIO-STATE.EDU
*Department of Computer and Information Science, The Ohio State University*
*Columbus, OH 43210 USA*

## Abstract

Visual thinking plays an important role in scientific reasoning. Based on the research in automating diverse reasoning tasks about dynamical systems, nonlinear controllers, kinematic mechanisms, and fluid motion, we have identified a style of visual thinking, *imagistic reasoning*. Imagistic reasoning organizes computations around image-like, analogue representations so that perceptual and symbolic operations can be brought to bear to infer structure and behavior. Programs incorporating imagistic reasoning have been shown to perform at an expert level in domains that defy current analytic or numerical methods.

We have developed a computational paradigm, *spatial aggregation*, to unify the description of a class of imagistic problem solvers. A program written in this paradigm has the following properties. It takes a continuous field and optional objective functions as input, and produces high-level descriptions of structure, behavior, or control actions. It computes a multi-layer of intermediate representations, called spatial aggregates, by forming equivalence classes and adjacency relations. It employs a small set of generic operators such as aggregation, classification, and localization to perform bidirectional mapping between the information-rich field and successively more abstract spatial aggregates. It uses a data structure, the *neighborhood graph*, as a common interface to modularize computations. To illustrate our theory, we describe the computational structure of three implemented problem solvers – KAM, MAPS, and HIPAIR — in terms of the spatial aggregation generic operators by mixing and matching a library of commonly used routines.

## 1. Introduction

It is commonly believed that there are two styles of scientific thinking: *analytical*, a logical chain of symbolic reasoning from premises to conclusions, and *visual*, the holding of imagistic, analogue representations of a problem in one's mind so that perceptual and symbolic operations can be brought to bear to make inferences. Neither style is to be preferred a priori over the other. However, for problems whose complexity precludes a direct analytical approach, a certain amount of qualitative and visual imagination is needed to provide the necessary "feel" or "understanding" of the physical phenomena. Once the picture is clear, the analytical mathematics can take over and lead more efficiently to logical conclusions. This "feel and physical understanding" is often considered to be informal, imprecise, and apparently unteachable, but necessary for scientists and engineers.

We believe part of this ability to visualize and imagine must consist of skills to generate images, discover structures and relations in the images, transform the structures, and predict how the structures respond to internal dynamics or external forcing.

While most AI work in visual reasoning has focused on diagrams and their role in controlling search, in recent years we have seen the development of a class of problem solvers that are *imagistic*, i.e., the problem solvers derive their power primarily from the use of visual apparatus and only secondarily from search and analytical methods. These problem solvers have been designed to perform tasks in many different domains: control and interpretation of numerical experiments (Yip, 1991; Nishida & et al., 1991; Zhao, 1994), kinematics analysis of mechanisms (Joskowicz & Sacks, 1991), design of controllers (Zhao, 1995; Bradley, 1992), analysis of seismic data (Junker & Braunschweug, 1995), and reasoning about fluid motion (Yip, 1995). However, there are important commonalities underlying them. In this paper, we present a framework to provide a unified description of this class of problem solvers. Our framework consists of three ideas:

- **The field ontology:** The input is a field, a mapping from one continuum to another. It is an image-like analogue representation. The field is assumed to have a metric so that it is meaningful to talk about closeness and continuity.[1]

- **Structure discovery:** A central problem to be solved is the transformation of the information-rich input to abstractions well-suited for concise structural and behavioral descriptions. The transformation can be thought of as successive mappings of the input space into more abstract spaces that hide details and group similar objects into equivalence classes.

- **Multi-layer spatial aggregates:** We propose (1) as representation the *neighborhood graph* to encode *explicitly* adjacency relations among objects at one level of abstraction, and (2) as building blocks of computational processes a small set of generic operators to construct, transform, classify, and search the neighborhood graph. The operators are recursively used to implement task-specific applications. The multi-layer theory has two advantages: (1) A nonlocal property of a lower layer can be redescribed as a local property of a higher layer, and (2) On each layer the neighborhood graph provides a common interface to support *identical* modular computations.

A field is a mapping from one continuum (say $R^m$) to another (say $R^n$). More concretely, one can visualize a m-dimensional space with a n-vector attached to each point in the space. Fields are commonplace in science and engineering applications. They are used to describe how physical quantities vary over space and time. Temperature in a room is a three-dimensional scalar field. Weather data can be described as a 4D-spacetime field with a 6-vector attached to each point: velocity (three components) of air flow, temperature (scalar), pressure (scalar), and density (scalar). Other examples of fields include the brightness intensity array in vision, the configuration space in mechanism analysis, and the phase space (vector field) of dynamical systems.

In actual computer representations, we often approximate a field with a grid. The grid may be uniform or non-uniform. The field can be reconstructed from numerical simulation or measurements. A field does not contain any symbolic abstractions; it is completely numerical. Fields are composable. One can extend the dimension of the underlying space and/or the number of components in the vector attached to each point of the space.

---

[1]. Forbus et al. (1991) proposed a general methodology for qualitative spatial reasoning: the Metric Diagram/Place Vocabulary (MD/PV). We generally agree with their methodology. Their paper inspired us to look for a more refined framework to unify a class of problem solvers that integrate visual and symbolic reasoning.

As a representation for physical systems, a field has two distinguishing characteristics. First, it is *information-rich* in the sense of the Shannon-Weaver measurement of information. An instantaneous field of a $128^3$-grid flow simulation may contain on the order of $10^8$ bits of information. Second, it is *pictorial* in the sense that structures and relations are only implicitly represented in the field.

As a consequence of both the information-richness and the pictorial quality, we argue that in reasoning about fields the *central computational problem* is the *efficient transformation of a pointwise field description of a physical system into economical symbolic abstractions well suited for explaining the structure and behavior of the system*.[2] Figure 1 illustrates how the field ontology relates to the other commonly used ontologies in Qualitative Physics: device (DeKleer & Brown, 1984), process (Forbus, 1984), and constraint (Kuipers, 1986). To be useful, the symbolic descriptions must impose a conceptual structure on the system so that the complexity of the system can be understood in terms of well-defined parts and subparts and interactions among them. The relevant parts and interactions are often *abstract* global properties of the field. An abstract property is a property whose support is large and nonlocal, whereas the support of a property is defined as the subset of a field on which the property depends. On the other hand, for computational complexity reasons we prefer to build the recognition procedures from basic routines that are local and independent of task-level information as much as possible. These considerations lead us to adopt an architecture where the pointwise description and the final symbolic descriptions are mediated by layers of equivalence classes of objects with *explicit* adjacency relations. We call such a layer of objects a *spatial aggregate*.

Where do spatial aggregates come from? In a real field, there tend to be continuities of properties (such as intensity or temperature or pressure) so that the field can be divided into equivalence classes, i.e., open regions where a particular property varies in an approximately uniform way. With continuities we can achieve an economy of description by focusing on the open regions and their boundaries instead of the pointwise field. Higher-order continuities, i.e., continuities of properties defined on the open regions, can similarly be used to build more abstract spatial aggregates.

The formation of equivalence classes presupposes the existence of continuity. This brings us to a methodological point. It is important to clearly identify the source of continuities in the field or equivalently in the physical system the field represents. The discovery of valid and general continuities in the physical system is as much a scientific contribution as the subsequent computational use of them to form an articulated conceptual model to explain structure and behavior.

Our motivation for this paper comes from the desire to understand the computational structures shared by a class of automatic problem solvers that integrate visual, symbolic, and numerical methods. We would like to make this computational structure explicit so that comparisons and generalizations can be made. Our goal is to develop a way of organizing programs around image-like analogue representations, and an appropriate language to make programs written in this style clear.

---

2. Inferring structural descriptions from a field can be an ill-posed problem (e.g., recovering 3D shapes from 2D images). To avoid these difficulties, we will assume the structure-recovery problem to be well-posed so that our main concerns are computational efficiency and appropriate abstractions.

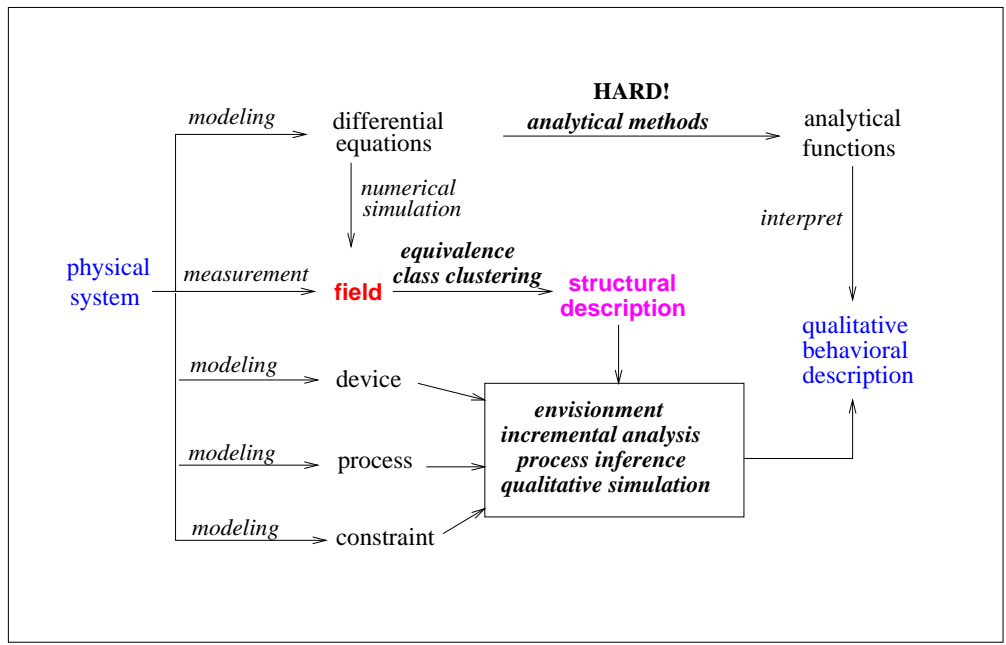

Figure 1: Field as an ontological abstraction for reasoning about physical systems. The diagram depicts the relationships among different ontologies used in Qualitative Physics. The central computational problem in field reasoning is the recovery of economical structural descriptions for qualitative behavior description and explanation. A key step in the structural recovery is the formation of equivalence classes. Identifying general and valid continuities on which useful equivalence class relations are based is an important scientific contribution.

The next section develops the theory of spatial aggregation in detail. Section 3 describes a language to support programs that are organized around neighborhood graphs. Section 4 illustrates the usefulness of the language by describing succinctly the computation structure of three implemented programs – KAM, MAPS, and HIPAIR. We choose these programs as illustrations largely because of our familiarity with them. Section 5 shows how to program in the spatial aggregation language, using an example from image analysis. We plan to investigate the applicability of our framework to several other programs, such as those constructed by Kuipers and Levitt (1988), Forbus et al. (1991), Gelsey (1995), and Junker and Braunschweug (1995).

## 2. Spatial Aggregation Theory

Given a field, a spectrum of reasoning tasks can be defined. The following list is roughly in the order of increasing complexity:

- **Infer structural descriptions.** Find out objects, if any, that exist in the field. What are their shapes, sizes, and locations? How are they distributed? How are they created? How do they evolve as some parameter (say time) is varied?

- **Classify.** Assign semantic labels to objects and configurations.

- **Infer correlations.** Determine how the geometry and distribution of one type of objects correlate with those of another type?

- **Check consistency.** Given two objects or configurations, test if they are equivalent or if they are pairwise consistent.

- **Infer incremental behavior.** Given an instantaneous configuration, predict its possible short-term behaviors.

- **Infer behavioral descriptions.** Explain and summarize the evolution of objects by a set of domain-specific interaction rules.

### 2.1 Requirements of imagistic reasoning

Partly motivated by Ullman's theory of visual analysis (Ullman, 1984), we find desirable the following general requirements on imagistic reasoning:

- **Abstractness.** The problem solver should be able to find objects defined by abstract global properties.

- **Open-endedness.** The problem solver architecture should be applicable to a variety of domains (fluid motion, seismic data, weather data, phase space, or configuration space). This requirement implies that the basic recognition routines must be modular and composable. Task-specific knowledge affects the choice and ordering of these routines.

- **Efficiency.** The "building blocks" of the recognition machinery must be local and non-goal-specific. "Non-goal-specific" means the operations of the building blocks do not depend on the interpretation of the objects they manipulate. This requirement implies that the basic routines should have local supports and in principle can run in parallel.

- **Soundness.** The structural and behavioral descriptions must be consistent with known physical and mathematical principles.

- **Succinctness.** The structural and behavioral descriptions should contain the qualitatively important distinctions relevant to the high-level tasks at hand.

## 2.2 Theory

Our theory of imagistic reasoning postulates the existence of multi-layers of spatial aggregates. Figure 2 shows the layers of spatial aggregates and computations organized around them. A primitive aggregate is defined as an equivalence class of subsets of the pointwise field representation. An aggregate is composed of equivalence classes of primitive aggregates. The field is assumed to have a task-dependent metric. The metric induces a topology on the space and hence it is meaningful to talk about adjacency. The data structure for a spatial aggregate is a *neighborhood graph* whose nodes represent objects and edges represent adjacency relations among the objects. The input field is sampled to form the lowest layer of abstraction; the field can also be affected by control actions from the higher-level abstraction layers.

Just as the *stream* construct in the SCHEME programming language provides a common interface for organizing signal processing computations, the neighborhood graph is our conceptual glue for piecing together operations that manipulate fields. We like to visualize nodes of neighborhood graph as open sets (in topology) in some appropriate space. Two nodes are *adjacent* if their respective open sets are contiguous.[3]

The topological notion of adjacency is amazingly useful in reasoning about physical systems. In grouping objects into equivalence classes, a cluster tends to give rise to a connected component of the neighborhood graph. In reasoning about kinematics, the neighborhood graph provides the essential connectivity information among free space regions. In finding "interesting" structures, the pairwise consistency of the adjacent nodes localizes search regions. In isolating bifurcation patterns, the mismatch of adjacent objects provides a hint for further analysis. In constraint propagation and path search, the adjacency structure imposes locality to increase computational efficiency. Prevalence and simplicity – these two aspects of the neighborhood graph make it a powerful data structure for unifying many spatial computations.

Our theory revolves around the computation of the neighborhood graph and the nature of the processes that construct, filter, transform, and compare neighborhood graphs. We isolate a set of generic operators *aggregate*, *classify*, *re-describe*, and *search* which correspond to the important conceptual pieces common to a class of imagistic problem solvers such as KAM (Yip, 1991), MAPS (Zhao, 1994), and HIPAIR (Joskowicz & Sacks, 1991).

The next section discusses these operators in detail. Section 4 illustrates the use of these operators in a rational reconstruction of three implemented computer programs.

## 3. The Language of Spatial Aggregation

We present a language for describing computational processes organized around spatial aggregates. The language provides a small set of operators to construct and manipulate neighborhood graphs. The operators make the conceptual structure of several implemented programs clear.

---

3. Let A and B be two open sets. A and B are contiguous if either $\bar{A} \cap B \neq \emptyset$ or $\bar{B} \cap A \neq \emptyset$ where $\bar{A}$ is the closure of the set A. In particular, if A and B overlap, then they are contiguous.

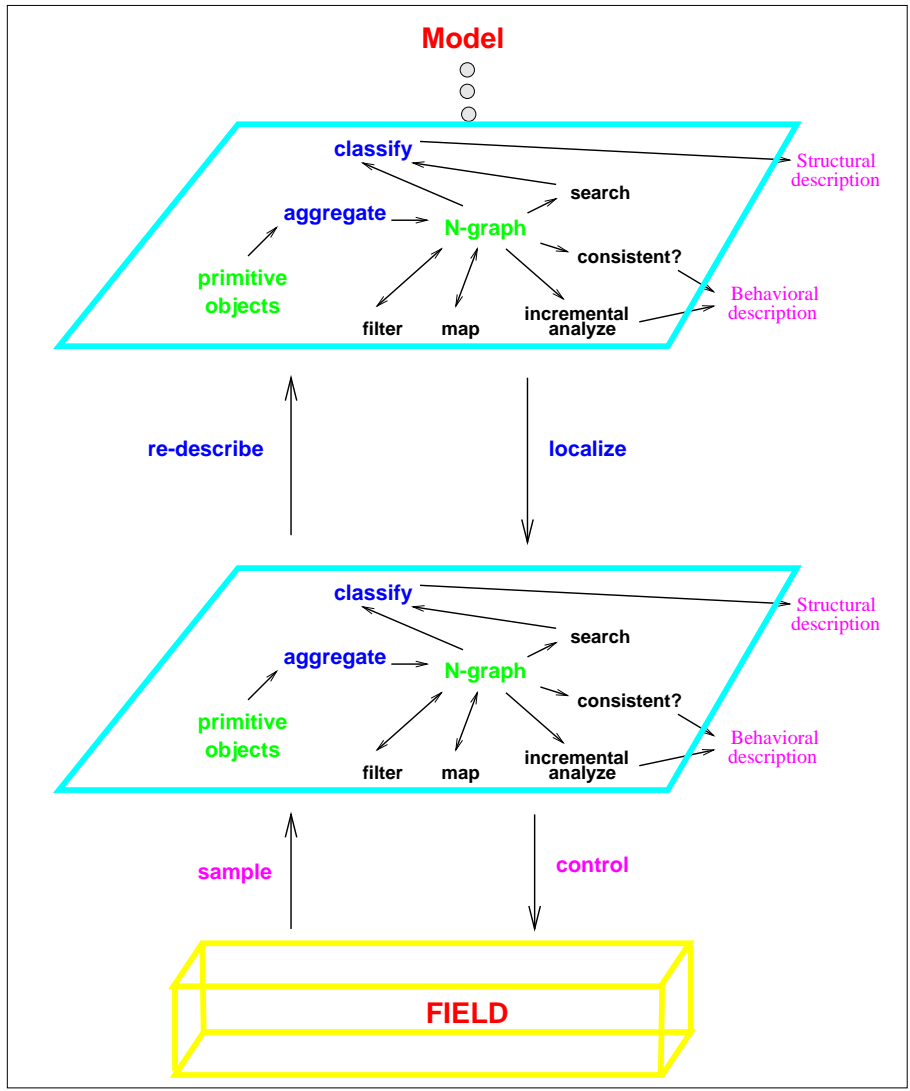

Figure 2: A schematic representation of the computational structure for analysis of a field ontology. There are multi-layers of spatial abstraction. An abstraction level is defined by the neighborhood graph, a data structure representing spatial aggregates and adjacency relations. The input field is fed to the lowest abstraction layer. Note the identical computational structure on each layer. The aggregate operator computes adjacency relations based on a task-specific metric. The neighborhood graph is the common interface for map and filter routines. The remaining operations correspond to the generic analysis tasks. A repertoire of task-independent geometric manipulation routines (which are not shown) are accessible by the generic operators.

### 3.1 Task-level operators

The task-level generic operators consist of `aggregate`, `classify`, `re-describe`, `localize`, `search`, `incremental-analyze`, together with the predicates `pairwise-consistent?` and `consistent?`. The neighborhood graph is the "conceptual glue": it allows the computation of hierarchical structural descriptions to be organized in a uniform manner. The following box summarizes what the language provides and what a user needs to supply in order to write programs in spatial aggregation.

---

**Language Features**

- User interface functions:

  `aggregate`, `classify`, `re-describe`, `localize`, `search`, `incremental-analyze`, `pairwise-consistent?`, `consistent?`

  *A user must specify the neighborhood relation, field metric, and equivalence relation for these operators.*

- Data types:

  - `N-graph` and its constructors, accessors, modifiers.
  
    Examples of `N-graph` include 4-adjacency arrays, minimal spanning tree, and Voronoi diagram.
  
  - Fields:
  
    bitmap, vector field, etc.

- Libraries:

  - Geometric utilities:
  
    `intrinsic-geometry`, `contain?`, `intersect`, $\partial$, $\delta$.
  
  - Numerical and image processing routines:
  
    FFT, convolution, integrator, linear system solver, vector/matrix algebra.

---

1. `aggregate(objects combiner)`

   The `aggregate` operator assembles a collection of objects into a spatial structure using the `combiner` procedure and explicates the spatial relations among the objects in terms of the neighborhood graph.[4] The operator returns a neighborhood graph (N-graph). The N-graph can be lazily built.

   For example, to recognize a trajectory in a phase space, the aggregate operator might be given a set of discrete points and a combiner procedure (such as minimal spanning tree) to establish adjacency relations. The combiner procedure might use a metric or topological properties of the underlying space.

2. `classify(N-graph cluster-proc class-rules)`

---

4. Recall the nodes in a neighborhood graph are objects and edges are adjacency relations.

The `classify` operator forms equivalence classes according to an equivalence relation (using the `cluster-proc`), and assigns a semantic label to each equivalence class — a subgraph of the input N-graph — according to the classification rules. For example, the orbit clustering procedure groups orbits into flow pipes.[5] The classification rules are a set of production rules. The operator returns a labeled N-graph.

The catalog of the classification labels is domain-specific. These classification labels serve as indices for storage and retrieval of shared class properties and methods for instantiating them.

3. `re-describe(N-graph desc-type)`

    The `re-describe` operator changes the representation of a primitive object. Like a *lambda* abstraction in SCHEME, this operator allows a compound object (say a subset of a N-graph) to be treated as a primitive.

    Given a classified object, the description-type procedure instantiates additional properties specific to that class of objects. For example, if a point set is classified as a space curve, it becomes sensible to compute additional geometric properties like length, curvature, and torsion.

4. `localize(N-graph select-proc enumerate-proc)`

    The `localize` operator systematically enumerates members of an equivalence class (nodes of N-graph) and selects those according to the select procedure. This operator "opens up" an abstraction to allow individual members of the equivalence class to be singled out.

5. `search(N-graph initial-states goal-p combiner)`

    The `search` operator returns paths starting from the initial-states and satisfying the `goal-p` predicate. The `combiner` procedure controls the order in which the graph is traversed.

6. `incremental-analyze(N-graph state-desc delta)`

    Given a N-graph and a description of states and constituent laws, the `incremental-analyze` operator computes the infinitesimal change to the qualitative state due to a small perturbation. The perturbation `delta` might be in the temporal, state, or parameter space.

There are predicates `pairwise-consistent?` and `consistent?`:

- `pairwise-consistent?(obj1 obj2 consistency-rules)`

    The `pairwise-consistent?` predicate decides if two objects are consistent according to the consistency-rules. The objects can be primitive objects such as nodes of an N-graph or N-graphs themselves.

- `consistent?(obj consistency-rules)`

    `Consistent?` tests if an object is well-formed according to the consistency-rules.

---

5. A flow pipe is a class of orbits that can be continuously deformed into each other. It is an example of the homotopy equivalence class.

### 3.2 Generic data structure and routines

The neighborhood graph is constructed by

- `N-graph-constructor(objects neighbor-p)`

  The N-graph-constructor takes a set of primitive objects and a neighborhood predicate as arguments, and returns a neighborhood graph. An example of such a neighborhood graph is the Voronoi diagram. The predicate neighbor-p tests if two nodes are neighbors.

The set of task-independent routines operate on the objects in the neighborhood graphs and support the task-level operations.

- `map(N-graph proc)`

  The map routine transforms a neighborhood graph using a prespecified procedure.

- `filter(N-graph mask)`

  A filter selects a subset of the neighborhood graph for further processing.

In addition to the generic operators, the language provides routines to perform common geometric manipulation. The following routines are especially useful:

1. `intrinsic-geometry(obj properties)` computes intrinsic geometric properties of objects (e.g., area, curvature, surface normal).

2. `contain?(obj1 obj2)` checks if obj2 is inside obj1.

3. `intersect(obj1 obj2)` computes intersection of two objects.

4. $\partial$`(object)` is the boundary operator that returns the boundary of an object. The dimension of boundary is co-dimension 1.

5. $\delta$`(object)` is the co-boundary operator that returns a new object whose boundary is the object. The dimension of the new object is one higher than that of the object.

6. `convolve(object mask)` performs pointwise convolution with the given mask.

### 4. Examples of Spatial Aggregation

In this section, we describe the architecture of three implemented systems KAM (Yip, 1991), MAPS (Zhao, 1994), and HIPAIR (Joskowicz & Sacks, 1991) in terms of the spatial aggregation framework. Although these programs are designed for different tasks, their computations share a strikingly similar pattern: These programs construct spatial objects, and interpret them via multi-layers of abstraction by object aggregation, classification, and re-description. Composite objects at a lower level are labeled and manipulated as primitive units at the next higher level.

Despite the fact that we are the authors of two of these programs, the structural similarities among these programs are not apparent to us until we carefully reconstructed these programs by defining the appropriate neighborhood graphs and generic operators. Analyzing these programs in a common framework will help us to understand not only what the programs do, but also greatly enhance our ability to construct future programs by a few spatial aggregation operators.

### 4.1 KAM

The task for KAM is to explore the dynamics of Hamiltonian systems and produce high-level summaries of their qualitative behaviors.

Given the state equations of a Hamiltonian system, KAM derives a symbolic description of its qualitative behavior — in terms of orbit types,[6] orbit bundles, phase portraits, and bifurcation patterns — from a collection of point sets representing orbits (or trajectories) in the phase space (see Figure 3). The point sets can be obtained from numerical simulation or measurements. To provide a useful interpretation of the point set, KAM has to decide (1) where to look for interesting orbits, and (2) how to group these orbits into larger structures. KAM proceeds via a sequence of intermediate representations that allow the gradual recovery of orbit structures and eventually the more global dynamical properties of the system. KAM is able to view an object at multiple levels of abstraction. For example, an orbit can be viewed as points in the phase space or a curve or part of an orbit bundle.

The computations in KAM are organized into four layers (as shown in Figure 4): (1) orbit, (2) orbit bundle, (3) phase portrait, and (4) bifurcation pattern. We will walk through the first level in sufficient detail to illustrate how the computation is synthesized from the spatial aggregation operators and neighborhood graph. Details of the remaining levels are described by Yip (1991).

The input is a point set. The `aggregate` operator imposes an adjacency relation on the point set by constructing a minimal spanning tree (MST). Two points are adjacent or neighbors if they are connected by an edge in the MST. Although the MST is appropriate for orbit interpretation, other applications might require different adjacency relations (such as Voronoi diagrams or k-nearest neighbors). The output of the `aggregate` operator is a neighborhood graph that encodes the edges of the MST.

The `consistent?` predicate checks if there are any inconsistent edges, i.e., edges that are significantly longer than their nearby edges, in the neighborhood graph. Deleting the inconsistent edge will partition the graph into subgraphs each of which represents a cluster of the original point set.

Next, the `classify` operator assigns a label, an orbit type, to the neighborhood graph according to the shape of the MST and the number of clusters. If the assignment is unsuccessful, KAM assumes the input point set does not contain enough points to reveal the structure of the orbit. KAM will request more points and repeat the aggregation step.

If the assignment is successful, the `re-describe` operator takes the labeled neighborhood graph and fills in information that is relevant to that particular orbit type. For example, if the orbit is a periodic orbit, the period of the orbit is determined. After filling in the details, the `re-describe` operator packages the orbit as a primitive object and passes it to the next level of abstraction, the orbit bundle level, where the same process of aggregation, consistency checks, classification, and re-description is repeated.

---

6. We introduce some useful terminology here. A dynamical system is a smooth vector field. An *orbit* is an integral curve of the vector field. An *orbit bundle* is a collection of adjacent orbits having the same qualitative behavior. A *phase portrait* is the collection of orbits that fill the phase space. A *bifurcation pattern* is a characteristic change in the structure of a phase portrait as some system parameters vary.

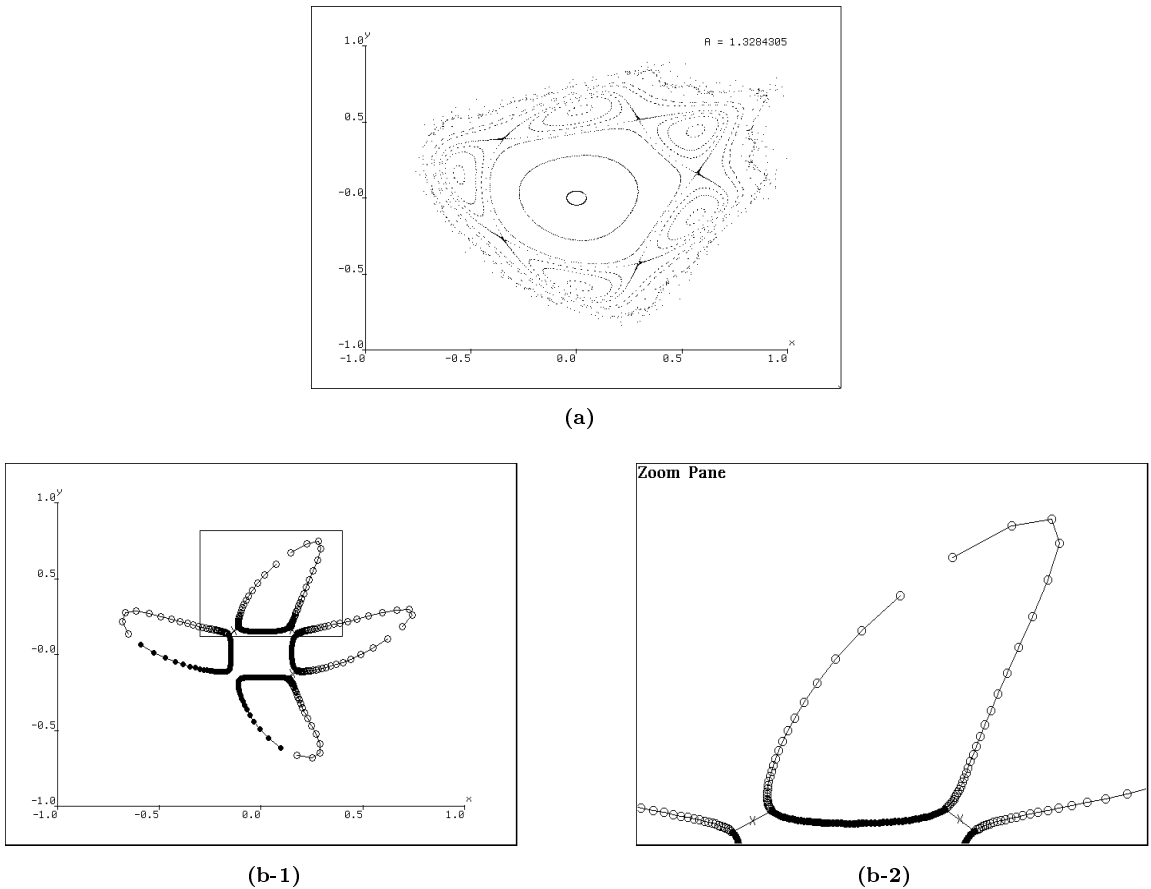

Figure 3: **Top:** (a) The phase portrait of a Hamiltonian system. The geometric structures in the phase portrait can vary drastically as the system parameter A changes. Like an expert dynamicist, KAM explores the dynamics of a nonlinear Hamiltonian system by finding interesting structures in the phase space. It decides what initial conditions and parameter values to try. It interprets what it finds and uses the structures it draws for itself to guide further exploration.
**Bottom:** (b-1) The minimal spanning tree representation of a point set. (b-2) Magnifying the boxed region — crosses ($\times$) are inconsistent edges. KAM imposes adjacency relations on a point set representing a trajectory in phase space. The structure of the minimal spanning tree reveals the type of the trajectory.

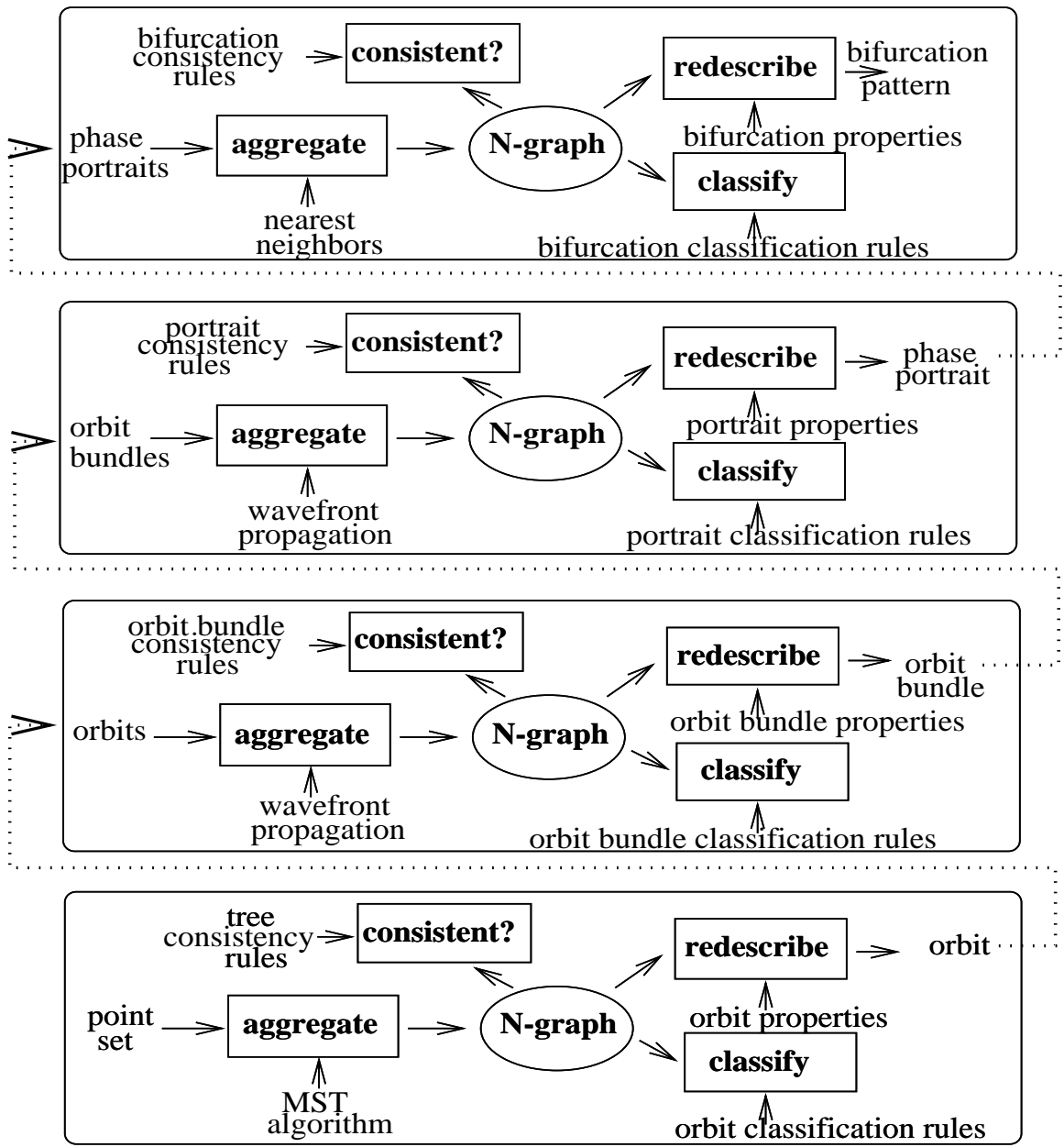

Figure 4: The computational structure of KAM viewed as spatial aggregation operators acting on neighborhood graphs. It has four layers of abstraction: orbit, orbit bundle, phase portrait, and bifurcation pattern. The computation is organized around neighborhood graphs. The structural similarities among the layers are apparent.

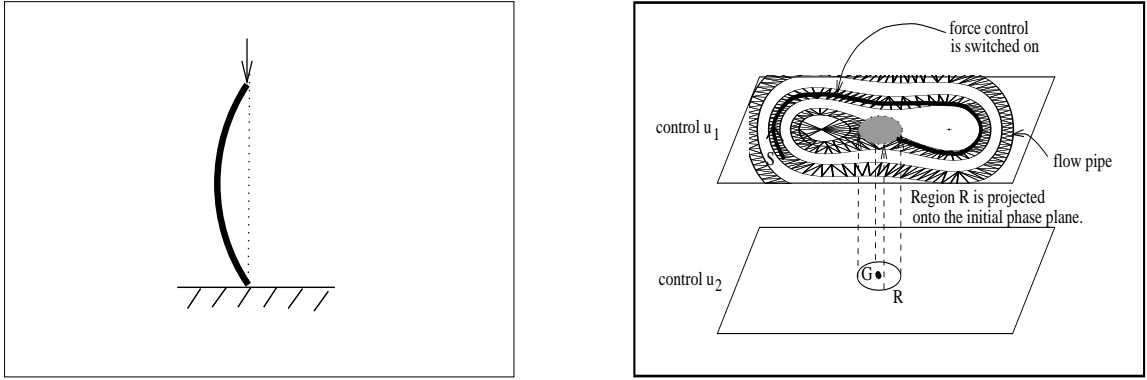

Figure 5: **Left:** Buckling of a beam due to an axial load.
**Right:** Phase spaces for the buckling beam (upper) and locally controlled beam (lower). To stabilize the buckling beam far from the unbuckled state — the unstable equilibrium G, MAPS (1) finds a flow pipe, a group of qualitatively similar trajectories, that reaches G, (2) deforms the trajectory emanating from the initial state via a force control until the trajectory is close to G, and then (3) switches to a conventional linear controller to achieve the desired stabilization. Let region R in the lower phase plane be a linearly controllable region with control $u_2$. Starting from an initial state S and initial control $u_1$, the system evolves along a trajectory within the flow pipe until it is close to the projection of the region R. The force control $u_1$ is turned on to deform the trajectory so that the system moves into the region R where a linear controller drives the system to the desired unbuckled state G.

### 4.2 MAPS

MAPS' task is to analyze the qualitative phase-space structures of dissipative systems and use the analysis results to guide the synthesis of control laws.

Like KAM, MAPS extracts high-level dynamical information from the phase space structures. But MAPS goes beyond KAM in two important aspects: (1) MAPS deals with three-dimensional structures explicitly (whereas KAM reasons with cross-sections of three-dimensional structures), and (2) MAPS uses the phase space structures to synthesize nonlinear control actions.

MAPS synthesizes a global control path geometrically (see Figure 5). Given an initial state and a desired state for the system under control, MAPS searches for a path in the phase space that connects the initial and the desired state. If the goal is not directly reachable from the initial state, MAPS pieces together multiple path segments by varying the control actions. A brute-force search for individual control paths in a continuum is clearly infeasible. MAPS partitions the continuous phase space into a manageable discrete set of objects — flow pipes — by defining appropriate equivalence relations, and searches out the flow pipes for good control paths.

The computations in MAPS are organized into four layers (as shown in Figure 6): (1) stability region, (2) flow pipe, (3) phase portrait, and (4) flow pipe graph. The input are the fixed points of the dynamical system[7]. Two fixed points are adjacent if they are connected to the same saddle by trajectories. The adjacency relation is represented by a neighborhood graph. The trajectories passing through the saddles are classified into equivalence classes and assigned stability region boundary labels. The `re-describe` operator computes the regions delimited by the stability region boundaries and represents them by polyhedra. The stability regions are fed to the next layer.

In the second layer, a stability region is triangulated by the Delaunay method. The `aggregate` operator constructs a neighborhood graph of the triangulation using the adjacency relation defined by the Voronoi diagram, the dual of the Delaunay triangulation. The triangulated sub-regions are classified into equivalence classes according to a topological criterion which states that two adjacent sub-regions are equivalent if the trajectories passing through them can be connected in a consistent manner. Equivalence classes of sub-regions are classified as flow pipes. Recall each flow pipe is a coarse representation of a set of trajectories having the same qualitative properties. The use of flow pipes simplifies considerably the control path planning problem.

The third layer aggregates the flow pipes to form a phase portrait.

The fourth layer is where control decisions are made. Flow pipes from *different* phase portraits are aggregated to form a larger structure, the flow pipe graph, which is the fundamental data structure supporting path planning in the phase space. Two flow pipes are adjacent if the phase space regions covered by the flow pipes overlap. Intuitively, one can switch from one flow pipe to an adjacent one by setting appropriate control parameters that generate the phase portraits in question. Given an initial and desired state, the `search` operator searches the flow pipe graph for solution paths.

Information can also be passed down the abstraction layer. Once a connected sequence of flow paths is found to satisfy a control objective, individual trajectory segments within the flow pipe are found by the `localize` operator using a shooting method.

### 4.3 HIPAIR

HIPAIR performs kinematic analysis of fixed-axes mechanisms built of rigid parts. Given a description of the shapes and motion types (such as translation and rotation) of the parts, HIPAIR derives realizable configurations of the mechanism.

HIPAIR derives realizable configurations of a mechanism by constructing and manipulating the *configuration space* of the mechanism (see Figure 7). The configuration space is the space of positions and orientations of the parts that make up the mechanism. HIPAIR partitions the configuration space into *free space regions* where parts do not overlap, and *blocked space regions* where they overlap. Only configurations that correspond to the free space regions are realizable. The boundaries of the free space regions are determined by the contact relations among the parts that touch each other. A *region diagram* is a graph whose nodes are free space regions and edges specify region adjacencies. The region diagram of the mechanism is composed of the regions diagrams of its pairwise interacting parts. For example, the region diagram of a mechanism with 10 parts is constructed from the region diagrams of 45 possibly interacting pairs.

---

7. Fixed points, or equilibrium points, are critical points in the phase space where the velocity vector vanishes. Fixed points are classified into three types according to the behavior of the nearby trajectories. A fixed point is an attractor if the nearby trajectories all move towards it. It is a repellor if they all move away from it. It is a saddle if some move towards and some move away from it.

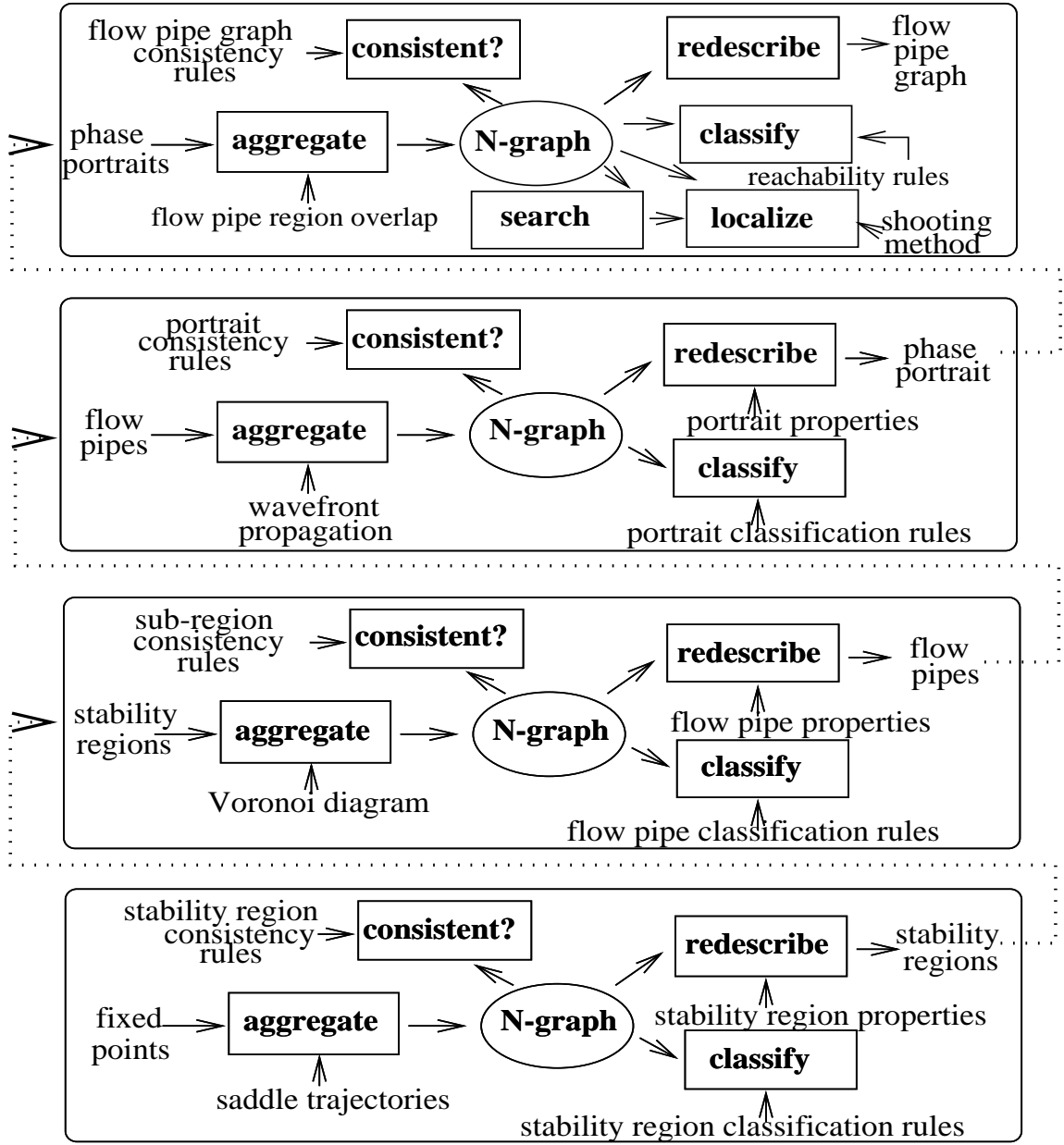

Figure 6: The computational structure of MAPS viewed as spatial aggregation operators acting on neighborhood graphs. It has four layers of abstraction: stability regions, flow pipes, phase portrait, and flow pipe graph. Note the structural similarities between KAM and MAPS. Control synthesis is implemented by the `search` and `localize` operators acting on the neighborhood graph representing the flow pipe graph.

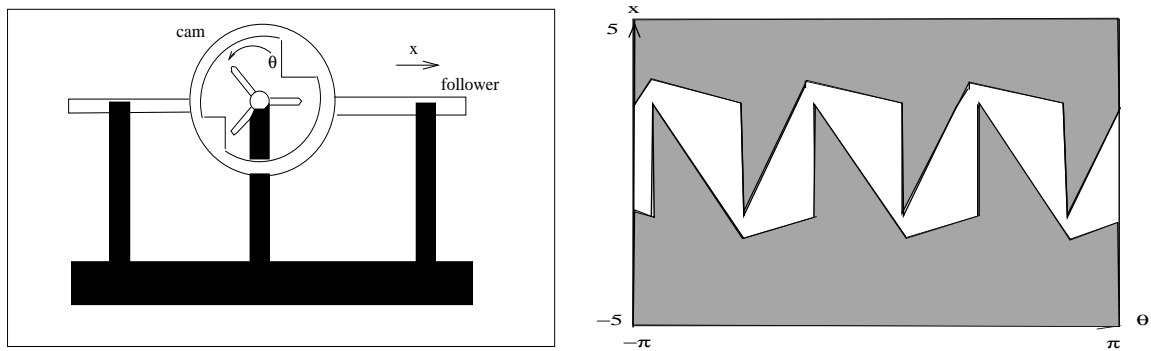

Figure 7: **Left:** The 3-finger cam-follower. **Right:** The configuration space for the cam-follower. $\theta$ is the cam rotation. $x$ is the follower displacement. The shaded regions are the blocked space, indicating that the parts overlap. The free space regions are the realizable configurations of the cam-follower. The boundaries of the free space regions are determined by the contact relations between the cam fingers and the follower.

The computations in HIPAIR are organized into three layers (as shown in Figure 8): (1) free space region, (2) subassembly region diagram, (3) mechanism region diagram. The input are the shapes of parts and their motion types. HIPAIR first considers a pair of interacting parts. It looks up the equations of the contact curves, i.e., curves in the configuration space for the pair corresponding to the configurations where the two parts touch, from a pre-compiled table of common contact curves. A contact curve is partitioned into segments by intersection points of the curve with either another contact curve or the boundaries of the configuration space. Two segments are adjacent if they share an endpoint. The `aggregate` operator assembles the segments and their adjacency relations into a neighborhood graph. The `search` operator traverses the neighborhood graph to find all closed chains of segments, where a closed chain of segments is a sequence of segments that intersect itself. Each closed chain of segment encloses a free space region. The `consistent?` predicate discards closed chains that lie inside other closed chains. The `classify` operator assigns a label to each closed chain, and the `re-describe` operator computes the free space regions delimited by the closed chains. Each free space region is subdivided into convex regions.

The input to the second layer are free space regions. They are aggregated into a neighborhood graph. Two free space regions are adjacent (or neighbors) if they touch. Given an initial configuration $S_0$ of an interacting pair, the `search` operator finds the free space regions reachable from $S_0$ by a depth first search. The neighborhood graph is re-described as a *subassembly region diagram*.

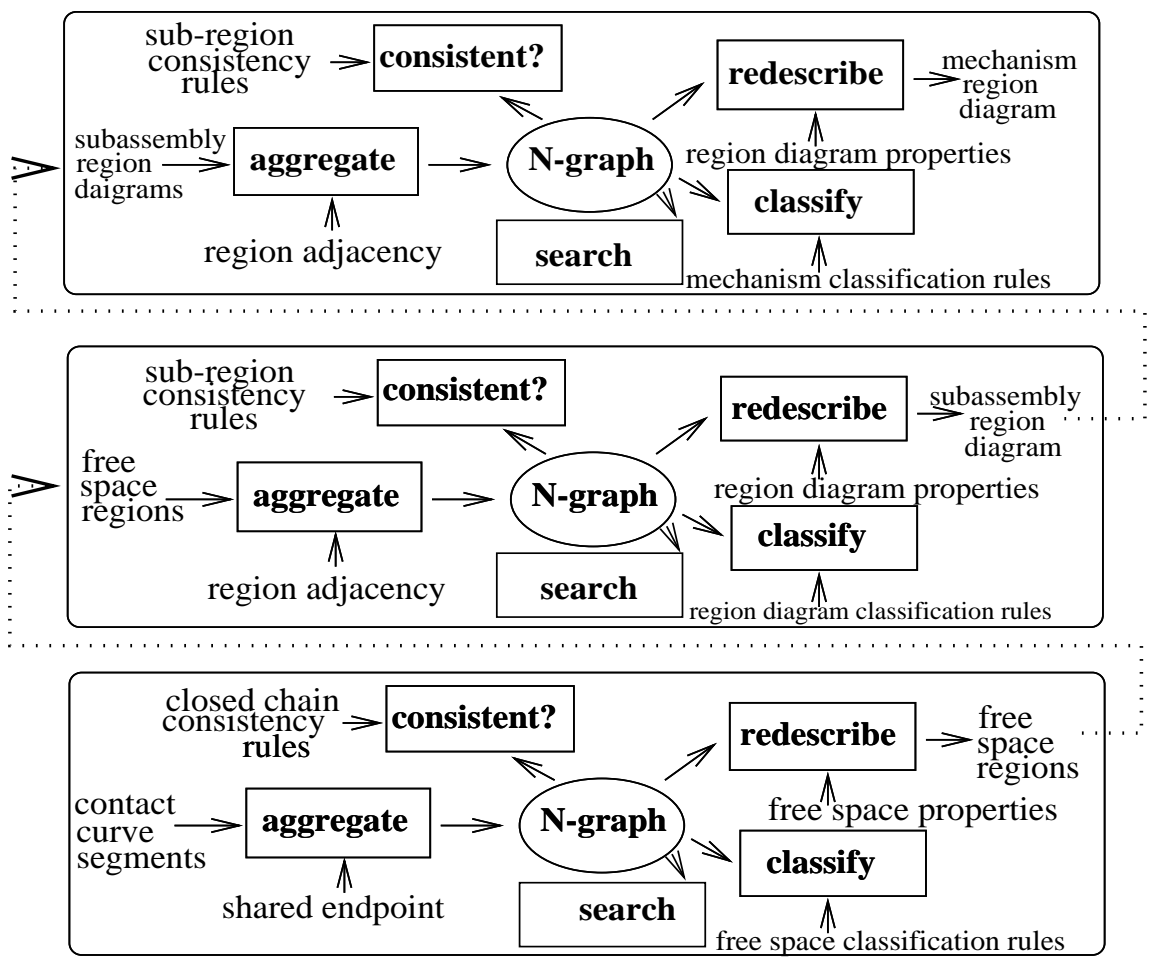

Figure 8: The computational structure of HIPAIR viewed as spatial aggregation operators acting on neighborhood graphs. It has three layers of abstraction: free space regions, subassembly region diagram, and mechanism region diagram. Note the structural similarities between HIPAIR, KAM, and MAPS. The `search` operator determines reachability conditions in all three layers.

On the third layer, HIPAIR combines all the subassembly region diagrams into a mechanism region diagram. The mechanism region diagram is a neighborhood graph whose nodes are realizable sets of free space regions and edges specify the adjacency of free space regions. A set of free space regions is realizable if their intersections are non-empty. For example, let $M_0 = \{R_0, S_0, T_0\}$ be a set of free space regions containing the initial configuration of a mechanism with three parts $P_1, P_2$, and $P_3$, where $R_0$, $S_0$, and $T_0$ are the free space regions in the subassembly region diagrams of the pairs $\{P_1, P_2\}, \{P_1, P_3\}$, and $\{P_2, P_3\}$ respectively. Suppose $R_0$ has one neighbor $R_1$, $S_0$ has one neighbor $S_1$, and $T_0$ has none. Then there are three candidate neighbors of $M_0$ given by:

$$M_1 = \{R_1, S_0, T_0\}$$

$$M_2 = \{R_0, S_1, T_0\}$$
$$M_3 = \{R_1, S_1, T_0\}$$

The `consistent?` predicate checks each of the candidate neighbors and discards the unrealizable ones.

## 5. An Illustration

In this section, we show what it is like to program in the spatial aggregation language. The example is a boundary tracer for line drawings.[8] We pick this example because image analysis routines can be quite naturally written in the spatial aggregation style.

Boundary tracing is a basic operation in image analysis.[9] The operation might be used to identify and group boundary segments from the same object. For example, consider a line drawing of overlapping 2D objects (see Figure 9). To group the boundary segments, one might first decompose the figure into segments, and junctions. A tracing process then joins colinear segments.

The input to the boundary tracing program is a bitmap:

```
0 0 0 0 0 0 0 0 0 0 0 0 0 0
0 1 1 1 1 1 1 0 0 0 0 0 0 0
0 1 0 0 0 0 1 0 0 0 0 0 0 0
0 1 0 0 0 0 1 0 0 0 0 0 0 0
0 1 0 1 1 1 1 1 1 1 1 1 1 0
0 1 0 1 0 0 1 0 0 0 0 0 1 0
0 1 0 1 0 0 1 0 0 0 0 0 1 0
0 1 1 1 1 1 1 0 0 0 0 0 1 0
0 0 0 1 0 0 0 0 0 0 0 0 1 0
0 0 0 1 0 0 0 0 0 0 0 0 1 0
0 0 0 1 0 0 0 0 0 0 0 0 1 0
0 0 0 1 1 1 1 1 1 1 1 1 1 0
0 0 0 0 0 0 0 0 0 0 0 0 0 0
```

The bitmap is rendered in Figure 10(a). Figure 11 illustrates how the output in Figure 10(b) and (c) is computed from the input bitmap, using the spatial aggregation operators.

We first define a neighborhood relation between pixels by the 4-adjacency (namely, the neighbors of a pixel are the pixels in its immediate north, east, south, and west). Because there is often no efficient way to construct N-graphs directly from neighborhood relations, we define an explicit N-graph neighborhood constructor that finds all the 4-adjacency neighbors of a given pixel.

Next the `aggregate` operator assembles the pixels into an N-graph by the N-graph constructor. Pixels in the N-graph are considered similar if they are neighbors and neither is a junction, where a junction is defined as a pixel whose value is one and which has more than two one-value neighbors. The `classify` operator groups the pixels into equivalence classes using a similarity threshold and returns the foreground equivalence classes, shown in Figure 10(b).

---

8. The details of the interpreter for the language, implemented in SCHEME, are discussed elsewhere (Bailey-Kellogg, Zhao, & Yip, 1996).
9. Jim Mahoney introduced us to a unified description of high-level operations on images.

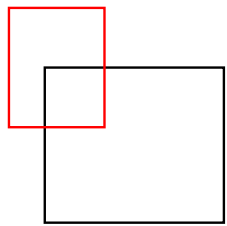

Figure 9: A line drawing of two overlapping objects.

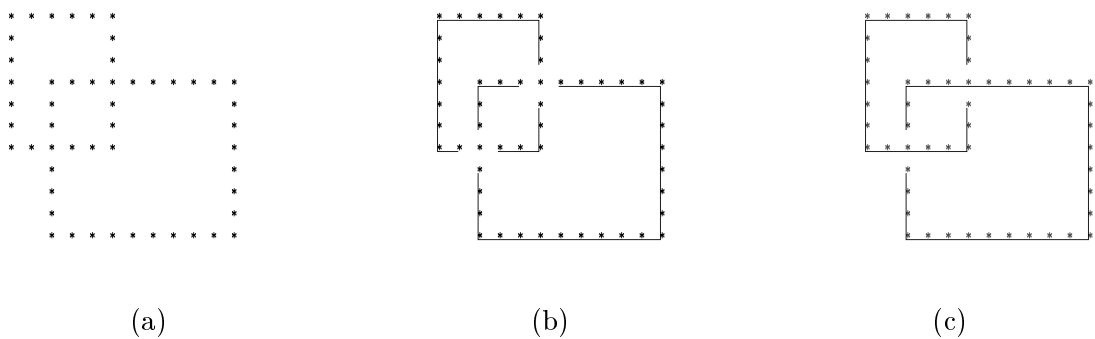

(a)  (b)  (c)

Figure 10: Boundary tracing operation in image analysis: (a) Pixels on boundaries of two overlapping objects; (b) Pixels are grouped into boundary segments; (b) Boundary segments are grouped into distinct object contours.

The foreground equivalence classes are then re-described as higher-level objects, boundary segments, which are in turn aggregated into a new N-graph using a different neighborhood relation. Specifically two boundary segments are neighbors if their minimum separation distance is less than a specified separation. Next, adjacent boundary segments which are colinear are grouped into equivalence classes, called contours. A contour represents the complete boundary of an object. Figure 10(c) shows the result of grouping.

We might want to check for impossible contours. A contour is legal if it is closed and not self-intersecting. Such conditions are expressed in a standard pattern language. Pairwise consistency rules can likewise be defined.

The program written in the spatial aggregation language is shown in Figure 12 and Figure 13.[10]

---

10. In the actual implementation of the language described by Kellogg, Zhao, and Yip (1996), the syntax of the operators differs slightly from those in Section 3.

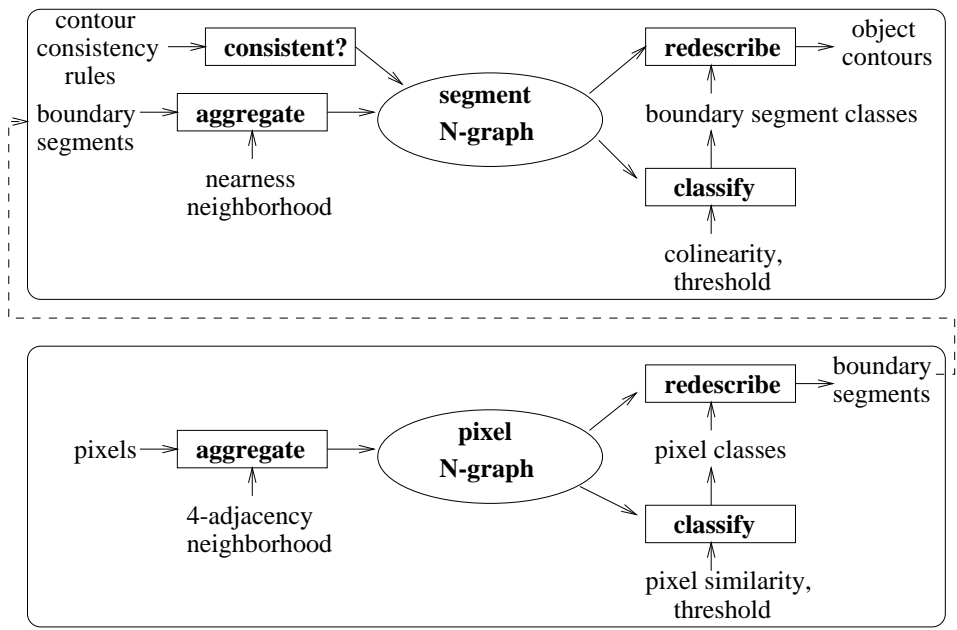

Figure 11: Boundary tracing operation: data flow in the spatial aggregation implementation.

## 6. Related Work

The literature in visual and spatial reasoning is enormous (e.g., Kosslyn, 1994; Glasgow, 1993). In this section, we discuss only the computationally oriented approaches.

The first line of work investigates how diagram-like representations aid heuristic search. Gelernter (1963) used diagrams in his geometry theorem prover to prune goals that are obviously false. Nevins' geometry theorem prover constrained forward deduction to conclude facts about objects explicitly depicted in the diagrams (Nevins, 1975). Stallman and Sussman (1977) exploited the connectivity and locality of lumped-parameter model to guide forward reasoning and implement symbolic constraint propagation. In a similar spirit, Larkin and Simon (1987) showed that in elementary mechanics problem a diagrammatic representation can reduce search because the diagram provides convenient indices for clustering objects and relations.

The second line of work concerns analogue simulations in naive physics. Funt's WHISPER program is the first AI program that uses primarily perceptual primitives to predict dynamical events in a simple blocks world (Funt, 1980). Arguing that the commonsense predictions of solid or fluid behavior cannot possibly depend on the solution of complicated equations, Gardin and Meltzer (1989) proposed a "molecular" simulation of strings and fluids. A body of fluid, for example, is decomposed into macro-molecules interacting with each other according to a small set of local rules. Chandrasekaran and Narayanan (1990) proposed a direct analogue simulation of the motion of a sliding block on an inclined plane. Their

```
;; 4-adjacency pixel neighborhood:
;;   neighbors are pixels one unit away using nearness ngraph
(define image-ngraph-fac
  (ngraph-near/instantiate image-field-fac 1))

;; Form a neighborhood graph for pixels
(define image-ngraph
  (aggregate pixels image-ngraph-fac))

;; Pixel classifier: two adjacent nodes are equivalent if they
;; have the same value and neither is a junction.
(define pixel/classify
  (classify-standard/instantiate
   image-ngraph-fac
   (lambda (n1 n2)
     (if (and (not (is-junction? n1))
              (not (is-junction? n2))
              (= (pixel/value n1) (pixel/value n2)))
         0 1))))

;; Form equivalence classes of foreground pixels
(define pixel-classes
  (filter
   (lambda (cl) (= (pixel/value (car cl)) 1))
   (pixel/classify image-ngraph pixels *threshold1*)))

;;; Form boundary segments
(define segments
  (redescribe pixel-classes segment/create))
```

Figure 12: Boundary tracing operation program (part 1): group pixels into boundary segments.

objective is to develop a cognitive architecture for visual perception and mental imagery. The direct representation of a scene they propose consists of a hierarchical, multi-resolution symbol structure encoding spatial relations among objects, and is linked to an analogical representation of the scene (image). The major challenge in analogue simulation is how to provide a reliable simulation without incorporating extensive physics and geometrical modeling.

The third line of work consists of spatial reasoning research in qualitative physics. Kuipers and Levitt (1988) described an approach to spatial reasoning in robot navigation and mapping of large-scale spaces. They proposed a four-level hierarchical representation incorporating topological and metric descriptions in terms of entities such as places, paths, distances, and angles. Forbus et al. (1991) developed the Metric Diagram/Place Vocabulary theory. The metric diagram contains both numerical and symbolic descriptions of a scene,

```
;; Boundary segment neighborhood defined by separation distance
(define segment-ngraph-fac
  (ngraph-near/instantiate segment-field-fac separation))

;; Form a neighborhood graph for boundary segments
(define segment-ngraph
  (aggregate segments segment-ngraph-fac))

;; Boundary segments classifier: two adjacent segments are
;; equivalent if they are colinear.  Two thresholds are used in
;; determining colinearity:  delta is the threshold for separation
;; distance between two end-points and epsilon is for the angle
;; between the tangent vectors at these end-points.
(define segment/classify
  (classify-standard/instantiate
   segment-ngraph-fac
   (lambda (s1 s2)
     (if (and (> (length (segment/points s1)) 1)
              (> (length (segment/points s2)) 1)
              (segment/colinear s1 s2 delta epsilon))
         0 1))))

;; Form contours, i.e., equivalence classes of boundary segments
(define segment-classes
  (segment/classify segment-ngraph segments *threshold2*))

;; Contour consistency check: closed and not self-intersecting
(define contour-consistency-rules
  '(if (and (closed? ?c)
            (not (self-intersecting? ?c)))
       #t #f))
```

Figure 13: Boundary tracing operation program (part 2): group boundary segments into distinct object contours.

while the place vocabulary is a quantization of the space according to task-specific criteria (see also footnote 1). Comparing the spatial aggregation framework and the MD/PV framework, we note two major differences. First, whereas a metric diagram is a mixed symbolic/quantitative representation, a field is purely numerical and does not encode any structures explicitly. Second, our theory postulates multi-layer spatial aggregates with identical computational structure at each layer. By focusing on the field ontology, which can be thought of as a special class of metric diagrams, we are able to emphasize the importance of the structure-recovery problem, and the commonalities underlying several implemented programs.

## 7. Conclusion

We have developed the spatial aggregation paradigm as a realization of imagistic reasoning. The paradigm systematizes the important task of interpreting time-varying information-rich fields. The paradigm consists of three ideas: (1) a field ontology, an image-like analogue representation, as input, (2) structural discovery – the efficient transformation from point-wise field representation to economical symbolic descriptions – as the central computational problem, and (3) a multi-layer neighborhood graph as the common interface and a small set of generic operators – aggregate, classify, redescribe, and search – as building blocks for computational processes that derive symbolic abstractions from the analogue representation. The paradigm relies on the important observations that the physical constraints on a real field (such as continuity and conservation) provide useful equivalence relations and economical descriptions, and a nonlocal property of a lower layer can often be redescribed as a local property of a higher layer.

The spatial aggregation paradigm supports the recovery of abstract properties via the multi-layer neighborhood graphs. It produces concise descriptions by manipulating equivalence classes of objects as primitives. It constructs modular programs from generic operators by mixing and matching a library of commonly used routines. It expresses task-specific knowledge in terms of field metric, adjacency relations, consistency predicates, classification rules, and redescription properties.

To illustrate our theory, we examine the computational structure of three implemented programs – KAM, MAPS, and HIPAIR – that integrate symbolic, numerical, and visual reasoning. We show a small set of generic operators that construct, transform, filter, classify, and search neighborhood graphs capture the commonalities of these programs. We develop a language, a way of organizing programs around neighborhood graphs, to make programs written in this style clear.

We are currently developing a toolkit to support problem solving using the generic operators of the spatial aggregation paradigm. Many research questions are still open. Can the operators be interfaced with computational geometry and with numerical analysis to build robust, efficient programs? What scientific problems can be solved by spatial aggregation?

Imagistic reasoning is a powerful strategy for mapping between analog signals generated by physical systems and discrete, symbolic representations of the systems. Spatial aggregation is only one of its many realizations. We believe that reasoning methods that derive their power primarily from perceptual operations on analog representations and only secondarily from search and analytical methods might prove effective in automating commonsense reasoning as well.

## Acknowledgements

We thank Chris Bailey-Kellogg for the help in implementing the spatial aggregation language, and the following people for helpful discussions and comments on the earlier drafts of this paper: Harold Abelson, Andy Berlin, B. Chandrasekaran, Gregor Kiczales, John Lamping, Shiou Loh, Jim Mahoney, Jeff May, Neal McDonald, Pandurang Nayak, Toyoaki Nishida, Elisha Sacks, Brian Smith, Jack Smith, Gerry Sussman, and Brian Williams.

KY is supported by an NSF National Young Investigator Award ECS-935777. FZ is supported by an NSF National Young Investigator Award CCR-9457802, an Alfred P. Sloan Foundation Research Fellowship, a grant from Xerox Palo Alto Research Center, a grant from AT&T Foundation, and an NSF grant CCR-9308639.